\pdfoutput=1

\documentclass[11pt]{article}

\usepackage[]{acl}

\usepackage{times}
\usepackage{latexsym}

\usepackage[T1]{fontenc}

\usepackage[utf8]{inputenc}

\usepackage{microtype}

\usepackage{inconsolata}

\usepackage{graphicx}
\usepackage{booktabs}
\usepackage{multirow}
\usepackage{algorithm}
\usepackage{algpseudocode}
\usepackage{amsmath}
\usepackage{caption}
\usepackage{subcaption}
\usepackage{hyperref}

\title{RECOST: External Knowledge Guided Data-efficient Instruction Tuning}

\author{Qi Zhang,~Yiming Zhang,~Haobo Wang,~Junbo Zhao \\
  Zhejiang University \\
  \texttt{\{cheung\_se,yimingz,wanghaobo,j.zhao\}@zju.edu.cn}}

\begin{document}
\maketitle
\begin{abstract}
In the current landscape of large language models (LLMs), the process of instruction tuning serves as an essential step. Considering the high computing power overhead, data-efficient instruction tuning was proposed to reduce the training data size in this process, aiming at selecting high-quality instructional data. 
Nevertheless, we argue that most current data-efficient instruction-tuning methods are highly dependent on the quality of the original instruction-tuning dataset. When it comes to datasets synthesized by LLMs, a common scenario in this field, dirty samples will even be selected with a higher probability than other samples.
To address these challenges, we utilized external knowledge (relevant examples or paragraphs) to evaluate those samples synthesized by LLMs with an in-context-based relative predictive entropy. 
Based on the new metric, we proposed a framework, dubbed as \textbf{RECOST}, which integrates external-knowledge-base re-ranking and diversity-consistent sampling into a single pipeline.
Through extensive experiments on several synthetic datasets (Alpaca and Alpaca-gpt4), we demonstrate the effectiveness of our method and achieve even better results with only \textbf{1\%} of the full dataset.
\end{abstract}


\section{Introduction}


Large Language Models (LLMs)~\cite{brown2020language} have demonstrated their remarkable capabilities in numerous fields of natural language processing (NLP) with the advancing of training datasets and the scale of model parameters. Behind this phenomenon, instruction tuning serves as an essential step to help pre-trained LLMs align to human cognition~\cite{ouyang2022training,peng2023instruction,chung2022scaling}. Instruction tuning refers to fine-tuning the LLMs on instruction-response pairs to endow LLMs with instruction-following capability and activate the knowledge gained in the pre-training period.

In the past two years, three types of instruction-following datasets have emerged~\cite{wang2023far}: those based on traditional NLP tasks (eg. Flan v2~\cite{weifinetuned,longpre2023flan}), those based on high-quality manual annotation (eg. LIMA~\cite{zhou2023lima}), and those synthesized by LLMs (eg. Alpaca~\cite{alpaca}). Among these, LIMA asserts that the quality of instruction-following datasets is far more important than their quantity. Thus, data-efficient instruction tuning is proposed to reduce the data size in instruction tuning without compromising the models' performance~\cite{zhou2023lima,chen2023maybe}.


Contemporary works in data-efficient instruction tuning predominantly concentrate on selecting high-quality data from instruction datasets synthesized by LLMs~\cite{Li2023FromQT, chen2023alpagasus, li2023one}.
Most approaches involve evaluation based on its proposed metrics, primarily centered around metrics related to predictive entropy~\cite{kadavath2022language}. 
We simply conclude these methods to synthetic-knowledge-guided ones as they select data points by using synthetic data points as prior knowledge.

\begin{figure}[h]
\centering
\includegraphics[width=\columnwidth]{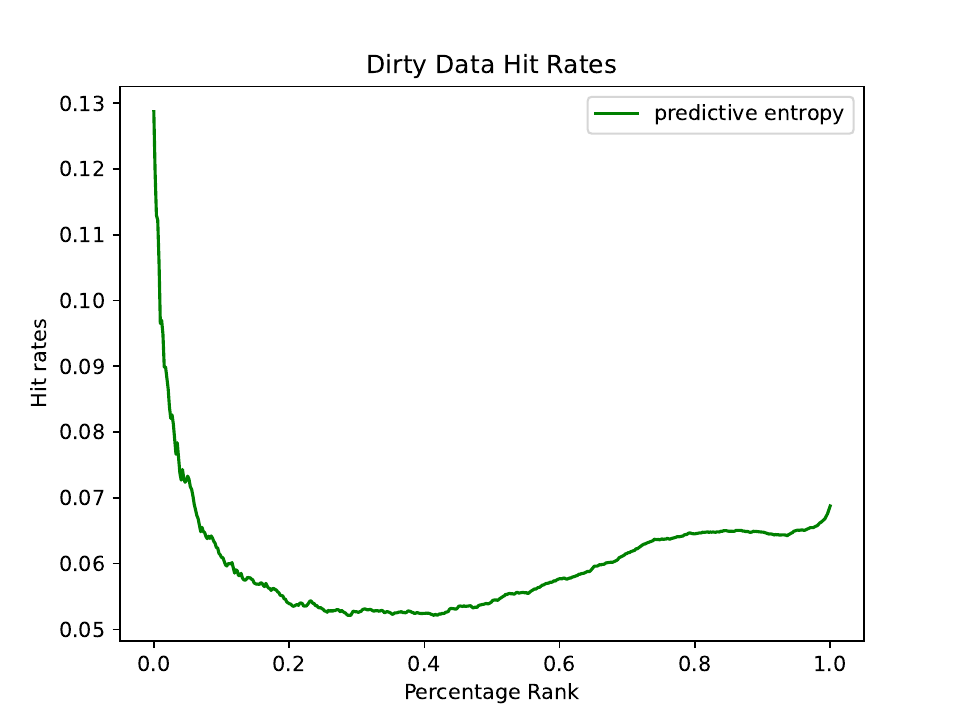}
\caption{The dirty data hit rates according to its predictive entropy calculated by LLaMA-2-7b. The horizontal axis represents the percentage ranking, while the vertical axis denotes the proportion of corrupted data within the data preceding that percentage threshold. Given the number of dirty data in the top $i$ data points as $d_i$, the hit rate at $i$ is calculated by $d_i/i$. The dirty data is collected by comparing Alpaca with Alpaca-cleaned.}
\label{fig:pe_dirty_hit_rates}
\end{figure}

However, ~\citeauthor{duan2023shifting} figured out that predictive entropy is not reliable enough. As shown in Figure~\ref{fig:pe_dirty_hit_rates}, \textbf{selection based on predictive entropy still results in the sampling of a non-negligible proportion of noisy data}. What’s even more surprising is that the higher the quality ranking, the greater the probability that samples contain noise, which is completely contrary to the original design intention. 
We argue that these methods heavily rely on the quality of the curated datasets which serve as the pre-experience~\cite{Li2023FromQT} or test set~\cite{li2023one} in their processes and are susceptible to the influence of outlier samples within the original datasets.
Therefore, in data selection under synthetic datasets scenarios, we argue that synthetic-knowledge-free methods are still under-explored. This issue currently imposes significant limitations on the development of the field of data-efficient instruction-tuning.

To address the challenge posed by the limitations of the predictive entropy in vanilla LLMs as outlined above, we utilize external information to evaluate samples synthesized by LLMs. 
Despite the suboptimal performance of this dataset in generative tasks~\cite{wang2023far}, its authenticity is significantly assured. 
But in the data-efficient instruction-tuning scenario of LLM, this cost is unacceptable.
Recognizing the importance of maintaining efficiency, 
we instead intuitively leverage pre-trained LLMs' intrinsic in-context learning (ICL) capabilities, treating these truthful samples as demonstrations.
Building on this foundation, we introduce a concept: in-context-knowledge-based relative predictive entropy, which serves as another dimension of uncertainty for vanilla LLMs.

In this paper, we propose \textbf{RECOST} (REtrieval, RE-rank, COreset sampling, and Supervised fine-Tuning), a framework that encompasses an in-context-knowledge-based re-ranking module and a diversity-consistent sampling module to avoid an overly homogeneous data distribution after re-ranking. 
With extensive experiments on synthetic datasets including Alpaca and Alpaca-gpt4, \textbf{RECOST} demonstrates its superiority over previous methods and surpasses remarkably the full-trained model with merely 1\% and 10\% training data on three benchmarks including the Alpagasus test sets~\cite{chen2023alpagasus,Li2023FromQT}, the OpenLLM benchmark~\cite{eval-harness} and AlpacaEval~\cite{alpaca_eval} benchmark.

All in all, our work explores how to instruct-tune LLMs under data-efficient scenarios with synthetic datasets. Our contributions can be summarized as follows:
\begin{itemize}
    \item We firstly propose \textbf{RECOST}, a method to mine high-quality data points from a synthetic dataset with consideration of the truthful-knowledge-based uncertainty and diversity.
    
    \item We conduct extensive experiments on multiple synthetic datasets and our method surpasses the fully trained model by utilizing only 1\% of the full dataset.

    
\end{itemize}

\begin{figure*}[h]
\centering
\includegraphics[width=0.9\textwidth]{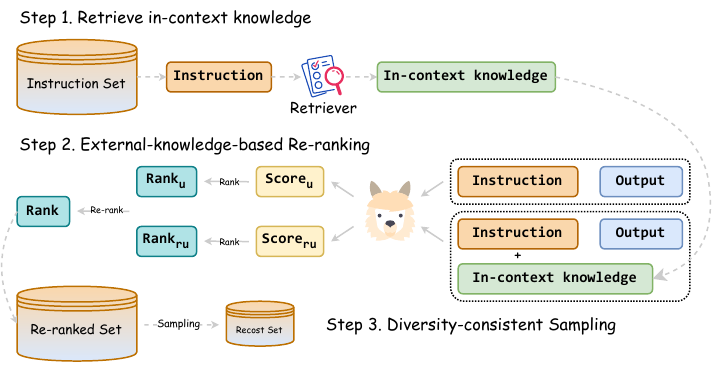} 
\caption{Overview of our proposed method. We start by retrieving in-context knowledge for each under-quantified data point. Two scores are produced by the vanilla LLaMA model on conditions with in-context knowledge or without that. The under-selected data points will be re-ranked by two ranks according to the produced two types of scores. Diversity-consistent sampling will be employed to select the qualified data points to finally supervised fine-tune the language models.}
\label{main_fig}
\end{figure*}


\section{Related Work}
\subsection{Instruction Tuning}
Instruction tuning has been regarded as an essential step to align pre-trained LLMs with human cognition~\cite{ouyang2022training,chung2022scaling,peng2023instruction}. This methodology refers to the supervised fine-tuning of pre-trained LLMs on datasets designed for instruction following. Each dataset entry comprises a pair, including an instruction and its corresponding response.

\subsection{Data-efficient Instruction Tuning}
As LIMA~\cite{zhou2023lima} makes the statement that \textit{less is more for alignment}, a new realm called data-efficient instruction tuning appears to help work out the bottleneck of data quality in this field. Recent works on data-efficient instruction tuning mainly quantify the quality of instruction data in two folds: the feedback from close-source LLMs and the score calculated by open-source LLMs based on a proposed metric.

Alpagasus~\cite{chen2023alpagasus} first dives into this field by employing ChatGPT to quantify the quality of instruction data by rating each data point, selecting those with higher scores, and supervised fine-tuning the LLMs with the data points with top scores. This work exploits a feasible way to sample high-quality data with feedback from close-sourced LLMs. Upon this, Deita~\cite{liu2024what} utilizes the feedback from ChatGPT to train scorer models to quantify the instruction data from the dimension of complexity and quality.

Another type of method, resembling the paradigm of active learning, tends to select data points based on the under-tuned model itself and an initial subset of the full dataset.
Instruction mining~\cite{cao2023instruction} summarizes the performance of several common metrics on data-efficient instruction tuning and proposes a complex equation to calculate the instruction data quality explicitly.
\citeauthor{Li2023FromQT} proposes Instruction-Following Difficulty (IFD), a self-guided method for mining data points with higher IFD scores. Nuggets~\cite{li2023one} introduces one-shot learning as implicit instruction tuning to guide the data selection for instruction tuning. Nuggets show promising results on some benchmarks while the selection process is computationally costly. Our method can also be regarded as one type of this but is more computationally efficient. 

\subsection{Synthetic Instruct-following Datasets}
Self-instruct~\cite{selfinstruct} serves as the milestone research in utilizing LLMs to synthesize instruct-following datasets. It starts from a small seed dataset and produces a fantastic instruction dataset with OpenAI's text-davinci-003. Upon this method, Alpaca~\cite{alpaca} was proposed by transferring self-instruct to ChatGPT. Alpaca-gpt4 shares the same prompts with the original Alpaca while using GPT4's response~\cite{peng2023instruction} as the answer to the prompts. Evol-instruct provides another paradigm for data synthesis by using ChatGPT to change the complexity of instructions and has generalized to multiple domains~\cite{xu2023wizardlm,luo2023wizardcoder,luo2023wizardmath}. In this paper, we primarily take synthetic instruct-following datasets into consideration.


\section{Methodology}
In this section, we will dive into the main methodology of \textbf{RECOST}. Figure \ref{main_fig} 
briefly illustrates the framework of our method.

\subsection{Preliminaries}
We start by defining the concept of predictive entropy. Predictive Entropy (PE), described in~\cite{kadavath2022language}, is a popular metric to measure the uncertainty of LLMs. It's defined as the entropy over the whole response $y$, which is equivalent to the accumulation of the token-wise entropy:
\begin{equation}
\label{eq:pe}
\begin{aligned}
    PE_{ic}(x,y)&=-\log p(y|x)\\
    &=\sum_i -\log p(z_i|y_{<i},x)
\end{aligned}
\end{equation}
where $x$, $y$, $z_i$ refer to the instruction, the response and the $i$-th token in $y$, respectively.

\subsection{Motivation and Methodological Prelude}
In this section, we will give an overview to elucidate the dilemma of current data-efficient instruction tuning and the insights of RECOST.

Most methods in data-efficient instruction tuning rely on building their metrics with predictive entropy. However, as highlighted by \citeauthor{duan2023shifting}, predictive entropy fails to adequately characterize LLMs' uncertainty. Besides, as shown in Figure~\ref{fig:pe_dirty_hit_rates} and Table~\ref{tab:hit_rates}, after sorting samples with their corresponding predictive entropy, the hit rate of dirty data has consistently maintained at an unacceptable level.

\begin{table}[h]
\resizebox{\columnwidth}{!}{%
\begin{tabular}{@{}c|ccccc@{}}
\toprule
Rank & $<$100    & $<$200    & $<$300    & $<$400    & $<$500    \\ \midrule
Hit Rates  & 12.9\% & 11.9\% & 11.3\% & 11.2\% & 10.6\% \\ \bottomrule
\end{tabular}%
}
\caption{The hits rates on dirty data when sorting samples with predictive entropy.}
\label{tab:hit_rates}
\end{table}

We hypothesize that this may be attributable to its excessively high correlation with response length, coupled with the tendency of LLMs to produce longer responses due to hallucination while synthesizing data.

Besides, since methods such as IFD~\cite{Li2023FromQT} introduce a subset of the synthetic dataset as the so-called pre-experience, the potential noise in sampled synthetic data points might further exert a negative impact on language models.

Thus, we come up with introducing external knowledge to propose a new metric and guide the data selection process. Originating from traditional NLP tasks, despite its poor performance in generative tasks~\cite{wang2023far}, Flan v2 is selected as an external knowledge source due to its low cost, vast data volume, and reliability. Considering efficiency, we refrain from fine-tuning the model on external knowledge; instead, we leverage the in-context learning capabilities of pre-trained LLMs to measure their external-knowledge-based conditional predictive entropy.
Recall the equation~\ref{eq:pe}, we here define it as in-context-knowledge-guided predictive entropy $PE_{ic}$ based on the in-context knowledge:
\begin{equation}
\begin{aligned}
    PE_{ic}(s,x,y)&=-\log p(y|x,s)\\
    &=\sum_i -\log p(z_i|y_{<i},x,s)
\end{aligned}
\end{equation}
where $s$ refers to the in-context knowledge. To further mitigate the randomness of demonstration selection, a retrieval technique is introduced for robustness.

Based on the foregoing, we will elaborate on the external-knowledge-based re-ranking module and the diversity-consistent sampling module of RECOST in the following sections.

\subsection{External-knowledge-based Re-ranking}
Previous research has shown that retrieved similar demonstrations can help improve LLMs' in-context learning ability~\cite{rubin2021learning,luo2024incontext,wang2024learning}. Intuitively, when it comes to synthetic samples, reliable data points will gain more from retrieved demonstrations than those less reliable ones. Therefore, based on predictive entropy and our proposed in-context-knowledge-guided predictive entropy, we further define the difference between the two entropies as the relative predictive entropy (RPE):
\begin{equation}
    PE_{r}(x,y)=PE(x,y)-PE_{ic}(s,x,y)
\end{equation}

On the one hand, as similar demonstrations can promote language models' in-context learning ability~\cite{luo2024incontext}, those data points with higher relative uncertainty can also be regarded as the more reliable ones. On the other hand, it's acknowledged that in-context learning is conducting an explicit gradient descent. From this perspective, a higher relative uncertainty signifies a greater sensitivity to external knowledge, thereby rendering such samples more amenable to learning.

Furthermore, to take both uncertainty and relative uncertainty into consideration, we define the mixed rank $R_m$ as the weighted average of the two ranks corresponding to the two types of uncertainty:
\begin{equation}
\label{eq:mixed_rank}
    R_{m}^{(i)}=w*R_{u}^{(i)}+(1-w)*R_{ru}^{(i)}
\end{equation}
where $R_{u}^{(i)}$ and $R_{ru}^{(i)}$ refer to the ranks of the $i$-th data point in the degree of uncertainty and relative uncertainty respectively.

At last, we re-rank all the data points with their corresponding mixed rank to generate the re-ranked instruction dataset.

\subsection{Diversity-consistent Sampling}
To enable the diversity of the sampled subset, we add an additional stage after re-ranking to further increase the diversity. Core-set sampling~\cite{sener2018active} is a technique for selecting a representative subset of a dataset, allowing for efficient approximation of solutions to problems by reducing computational complexity without significantly compromising result quality.

However, integrating greedy core-set sampling into our framework directly fails to make itself aware of the mixed ranks in Equation~\ref{eq:mixed_rank}. Thus, in this section, we introduce a sliding window mechanism, which is illustrated in Algorithm \ref{alg:coreset}, to consider the diversity of the sampled subset with awareness of its overall uncertainty.

\begin{algorithm}[htbp]
\caption{Core-set sampling with a sliding window.}
\label{alg:coreset}
\textbf{Input}: The re-ranked instruction dataset $D$, the sampling size $s$, the initial subset size $s_i$, the sliding window size $w$, and the tolerance $t$.

\textbf{Output}: The sampled subset $D_{s}$.

\begin{algorithmic}[1] 
\State $D_s\gets D[:s_i]$
\State $W\gets D[s_i:s_i+w]$
\For{$j\gets 1$ \textbf{ to } $w$}
    \State $T[j]\gets t$
\EndFor
\For{$k\gets 1$ \textbf{ to } $s-s_i$}
    \State $d\gets \mathbf{FarthestFirst}(D_s,W)$
    \State $D_s\gets D_s\cup W[d]$
    \State $W.\textbf{pop}(d);T.\textbf{pop}(d)$
    \State $\mathbf{update}(W,T)$
    \State $W.\textbf{push}(d);T.\textbf{push}(t)$
\EndFor
\State \textbf{return} $D_s$
\end{algorithmic}
\end{algorithm}

We start by sampling the top data points in the re-ranked instruction as an initial set. Similar to conventional core-set sampling, our algorithm also samples data points iteratively. However, within each iteration, a sliding window mechanism is introduced to enable the algorithm to only sample from the under-selected data points with higher uncertainty and relative uncertainty. Moreover, to further increase the diversity of the sampled subset, we introduce a tolerance $t$ for each data point in the sliding window to make sure each of them can only be considered for $t$ times at most. After every iteration, data points with zero tolerance will be erased from the sliding window.




\section{Experiments Settings}
\subsection{Datasets}
Self-instruct~\cite{selfinstruct} is a milestone method for constructing instruction-tuning datasets by distilling from closed-sourced LLMs. The Alpaca~\cite{alpaca} dataset employed self-instruct method to distill instruction data from ChatGPT or GPT-4~\cite{cao2023instruction} and serves as a commonly used dataset for instruction tuning. In this paper, we start with ChatGPT and GPT-4 versions of Alpaca.

\subsection{Benchmarks}
\label{sec:eval}
For evaluation, we use three benchmarks to evaluate our method.

For the general ability of instruction-tuned language models, we use the OpenLLM benchmark\footnote{\url{https://huggingface.co/spaces/HuggingFaceH4/open_llm_leaderboard}}, which includes four datasets: Arc~\cite{Yadav_2019}, Hellaswag~\cite{zellers2019hellaswag}, MMLU~\cite{hendrycks2021measuring}, and Truthfulqa~\cite{lin2022truthfulqa}. Following the general setting, we use 25-shot for Arc dataset, 10-shot for the Hellaswag dataset, 5-shot for the MMLU dataset, and 0-shot for the Truthfulqa dataset.

Another way to evaluate the instruction-tuned language models is to use LLMs to evaluate the responses generated by those models. We use the AlpacaEval\footnote{\url{https://github.com/tatsu-lab/alpaca_eval}} benchmark~\cite{alpaca_eval} to evaluate the open-end generation ability. To be specific, we use GPT4-turbo to choose the preferred response generated by our fine-tuned models and OpenAI's text-davinci-003 for a given instruction. The overall win rate will be devoted to evaluating the generation ability of language models.

\begin{table*}[h]
\centering
\resizebox{0.98\textwidth}{!}{%
\begin{tabular}{@{}cccccccc@{}}
\toprule
\multicolumn{1}{c|}{\multirow{2}{*}{Method}} & \multicolumn{1}{c|}{\multirow{2}{*}{\textbf{Data size}}} & \multicolumn{5}{c|}{OpenLLM   Benchmark} & \multirow{2}{*}{\textbf{AlpacaEval}} \\
\multicolumn{1}{c|}{}                     & \multicolumn{1}{c|}{}      & Arc   & Hellaswag & MMLU  & Truthfulqa & \multicolumn{1}{c|}{\textbf{Avg}}   &       \\ \midrule
\multicolumn{1}{c|}{LIMA~\cite{zhou2023lima}}            & \multicolumn{1}{c|}{1000}  & 55.55 & 81.55     & 47.74 & 47.23      & \multicolumn{1}{c|}{58.02} & 26.58 \\\midrule
\multicolumn{8}{c}{Alpaca Results}                                                                                                                   \\ \midrule
\multicolumn{1}{c|}{Full Alpaca}          & \multicolumn{1}{c|}{52002 (100\%)} & 54.18 & 78.21     & 45.80 & 42.05      & \multicolumn{1}{c|}{\emph{55.06}} & \emph{27.75} \\
\multicolumn{1}{c|}{IFD~\cite{Li2023FromQT}}                  & \multicolumn{1}{c|}{3111 (6\%)}  & 57.94 & 80.37     & 44.19 & 40.62      & \multicolumn{1}{c|}{55.78} & 36.78 \\
\multicolumn{1}{c|}{Random*}   & \multicolumn{1}{c|}{\textbf{520 (1\%)}}   & 54.10 & 78.22     & 47.52 & 39.77      & \multicolumn{1}{c|}{54.90} & 26.52 \\
\multicolumn{1}{c|}{Predictive Entropy*}            & \multicolumn{1}{c|}{\textbf{520 (1\%)}}  & 55.03 & 77.20     & 45.18 & 43.84      & \multicolumn{1}{c|}{55.31} & 35.59 \\
\multicolumn{1}{c|}{\textbf{RECOST} (ours)}   & \multicolumn{1}{c|}{\textbf{520 (1\%)}}   & 56.48 & 77.73     & 45.80 & 44.27      & \multicolumn{1}{c|}{\textbf{56.07}} & \textbf{39.19} \\ \midrule
\multicolumn{8}{c}{Alpaca-gpt4 Results}                                                                                                              \\ \midrule
\multicolumn{1}{c|}{Full Alpaca-gpt4}          & \multicolumn{1}{c|}{52002 (100\%)} & 56.57 & 80.72     & 49.06 & 54.51      & \multicolumn{1}{c|}{\emph{60.21}} & \emph{61.80} \\
\multicolumn{1}{c|}{Random*}   & \multicolumn{1}{c|}{\textbf{5200 (10\%)}}  & 55.63 & 80.87     & 48.52 & 51.27      & \multicolumn{1}{c|}{59.07} & 55.92 \\
\multicolumn{1}{c|}{Predictive Entropy*}            & \multicolumn{1}{c|}{\textbf{5200 (10\%)}}  & 57.59 & 81.19     & 47.95 & 52.13      & \multicolumn{1}{c|}{59.72} & 60.39 \\
\multicolumn{1}{c|}{\textbf{RECOST} (ours)}   & \multicolumn{1}{c|}{\textbf{5200 (10\%)}}  & 57.68 & 80.63     & 48.53 & 52.11      & \multicolumn{1}{c|}{\textbf{59.74}} & \textbf{63.35} \\ \bottomrule
\end{tabular}
}
\caption{Performance of RECOST on the OpenLLM Benchmark and AlpacaEval. The methods marked with * are the data-efficient instruction-tuning results we implemented based on the corresponding metrics, for better comparison with our metric. RECOST outperforms all previous methods on AlpacaEval and is comparable on OpenLLM Benchmark with a fraction of training data of them.}
\label{tab:main_results}
\end{table*}

Moreover, we follow previous research with similar settings and evaluate our proposed method with fully trained models Alpagasus test sets (Vicuna, Koala, WizardLM, and Self-Instruct) and IFD's additional test set (LIMA). In detail, we use GPT4 to rate two responses from two models on a scale of 1 to 10, which implies accuracy and relevance. To dismiss the potential positional bias, we also follow the 'Win-Tie-Lose` rule to judge the two responses both obversely and reversely:
\begin{itemize}
    \item \textbf{Wins}: RECOST wins twice, or wins once and draws once.
    \item \textbf{Ties}: RECOST draws twice, or wins once and loses once.
    \item \textbf{Loses}: RECOST loses twice, or loses once and draws once.
\end{itemize}

\subsection{Implementation Details}
For the retrieving period, we use a subset of Flan v2 as the knowledge source with consideration of time efficiency, which includes 10 million samples. Following the settings of \citeauthor{llm_embedder}, we use llm-embedder as our retriever and retrieve 5 related demonstrations for each data point.

We choose LLaMa-2-7b as the base model to validate our proposed method. All the models are trained with the Adam optimizer with a batch size of 64 and a 2e-5 learning rate for 3 epochs as official Alpaca.

\section{Experiments Results and Analysis}
\label{sec:exp}
In this section, we present the main results of our method on the three benchmarks mentioned in Section~\ref{sec:eval}. Moreover, we conduct extensive ablation studies based on several possible factors.

\begin{figure}[h]
\centering
\begin{subfigure}[b]{\columnwidth}
\centering
\includegraphics[width=\columnwidth]{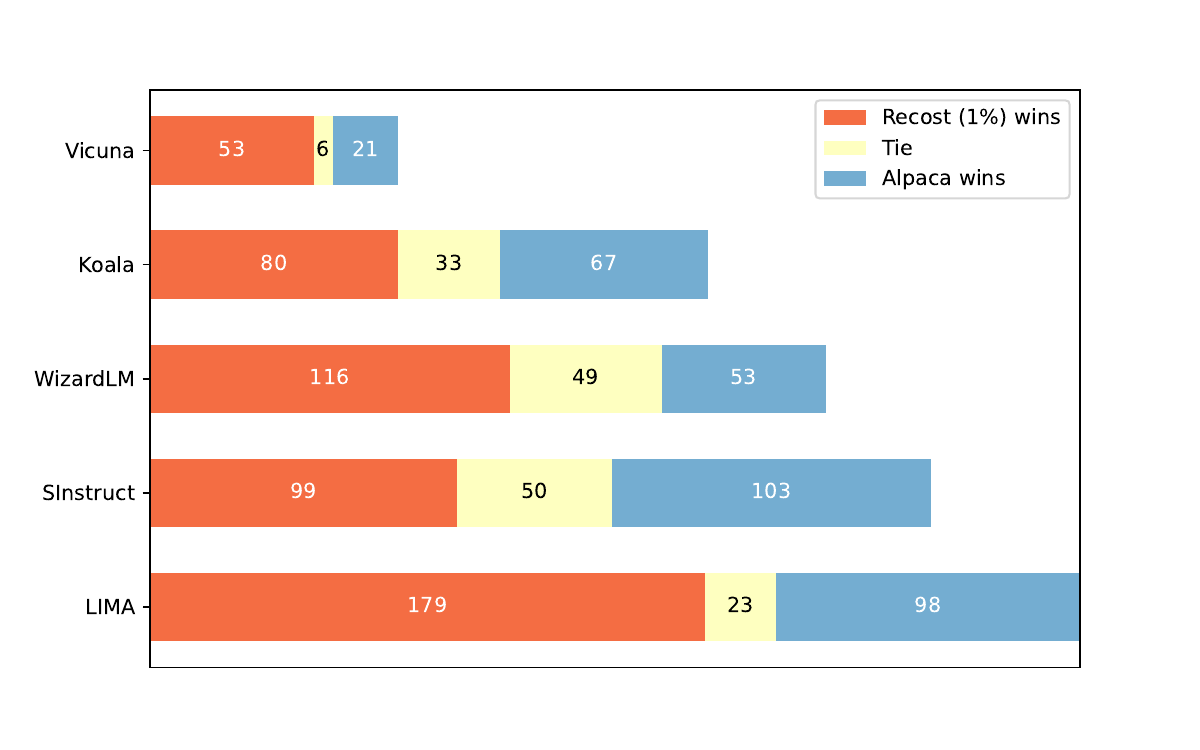}
\caption{RECOST vs. Full Alpaca.}
\label{fig:recost}
\end{subfigure}
\begin{subfigure}[b]{\columnwidth}
\centering
\includegraphics[width=\columnwidth]{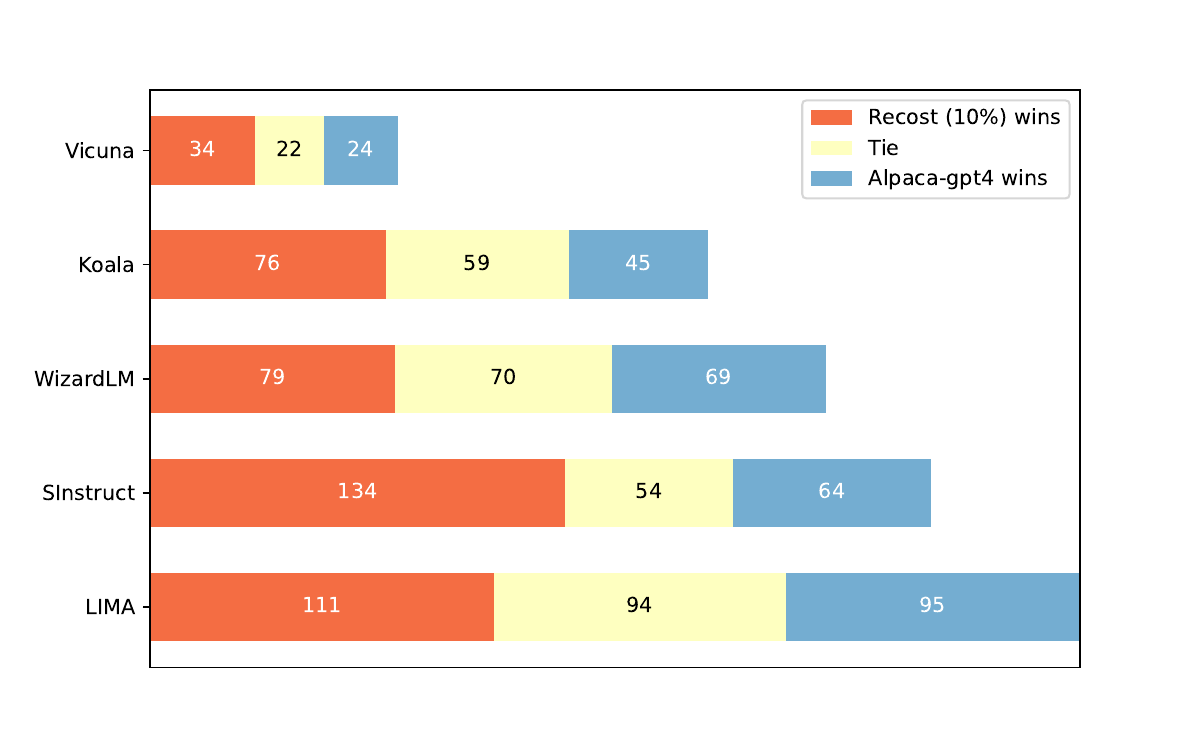}
\caption{RECOST vs. Full Alpaca-gpt4.}
\label{fig:recost-gpt4}
\end{subfigure}
\caption{Results on Alpagasus Test Set. Figure~\ref{fig:recost} and Figure~\ref{fig:recost-gpt4} demonstrate RECOST's performance compared to models fully trained on Alpaca and Alpaca-gpt4 respectively.}
\label{fig:recost-vs-alpaca}
\end{figure}

\subsection{OpenLLM Benchmark and AlpacaEval Benchmark}
The experimental results of OpenLLM Benchmark and AlpacaEval Benchmark are presented in Table~\ref{tab:main_results}.

Concerning the OpenLLM Benchmark, attributable to its constrained output domain, abbreviated output duration, and the assurance of a stochastic interval, our methodology realizes a marginal enhancement. In the context of AlpacaEval, as it constitutes an open-ended generation endeavor, our strategy markedly surpasses alternative methodologies by supplying data of superior quality.

Specifically, for the Alpaca dataset, RECOST outperforms remarkably the fully trained Alpaca model and random selection on both benchmarks. Besides, it surpasses the previous method with even less data. As for the Alpaca-gpt4 dataset, which is of superior quality, RECOST still achieves comparable outcomes to those of full training while only utilizing 10\% of the dataset volume, significantly surpassing the results of random sampling.

\subsection{Alpagasus Test Set}
Following Alpagasus~\cite{chen2023alpagasus} and IFD~\cite{Li2023FromQT}, we further compare our method with fully trained models on Alpagasus test sets (Vicuna, Koala, WizardLM, and Self-Instruct) and IFD's additional test set (LIMA). Figure~\ref{fig:recost} gives 


As shown in Figure~\ref{fig:recost-vs-alpaca}, RECOST shows promising results over the fully trained models on both datasets and only lags on one of the test sets on the Alpaca dataset.

\subsection{Ablation Study}
To evaluate the robustness of our method, we conduct extensive ablation experiments based on the following factors.

\subsubsection{Ablation on Knowledge Source}
\paragraph{Different External Knowledge Sources}
Besides Flan v2, inspired by retrieval-based generation (RAG), we also explore our method with knowledge from raw text. Following the settings of Self-RAG, we retrieve related paragraphs with the under-selected instruction data. To be specific, we also retrieve five paragraphs as in-context knowledge for every data point. As shown in Table \ref{tab:knowledge_source}, \textbf{RECOST} with Wikipedia as external knowledge also achieves promising results on both datasets. However, due to the lack of RAG capability for vanilla LLMs, RECOST (Wiki) is slightly inferior to RECOST (Flan v2) but still comparable with previous work.

\begin{table}[h]
\centering
\resizebox{\columnwidth}{!}{%
\begin{tabular}{@{}cccc@{}}
\toprule
\multicolumn{1}{c|}{\multirow{2}{*}{Method}} &
  \multicolumn{1}{c|}{\multirow{2}{*}{Data size}} &
  \multicolumn{1}{c|}{\multirow{2}{*}{OpenLLM}} &
  \multirow{2}{*}{AlpacaEval} \\
\multicolumn{1}{c|}{}                     & \multicolumn{1}{c|}{}      & \multicolumn{1}{c|}{}      &       \\ \midrule
\multicolumn{4}{c}{Alpaca Results}                                                                          \\ \midrule
\multicolumn{1}{c|}{Full Alpaca}          & \multicolumn{1}{c|}{52002} & \multicolumn{1}{c|}{55.06} & 27.75 \\
\multicolumn{1}{c|}{RECOST   (Wiki)} & \multicolumn{1}{c|}{520}   & \multicolumn{1}{c|}{55.51} & 37.76 \\
\multicolumn{1}{c|}{RECOST (Flan   v2)}   & \multicolumn{1}{c|}{520}   & \multicolumn{1}{c|}{\textbf{56.07}} & \textbf{39.19} \\ \midrule
\multicolumn{4}{c}{Alpaca-gpt4 Results}                                                                     \\ \midrule
\multicolumn{1}{c|}{Full Alpaca-gpt4}          & \multicolumn{1}{c|}{52002} & \multicolumn{1}{c|}{\textbf{60.21}} & 61.80 \\
\multicolumn{1}{c|}{RECOST   (Wiki)} & \multicolumn{1}{c|}{5200}  & \multicolumn{1}{c|}{59.78} & 61.61 \\
\multicolumn{1}{c|}{RECOST (Flan   v2)}   & \multicolumn{1}{c|}{5200}  & \multicolumn{1}{c|}{59.74} & \textbf{63.35} \\ \bottomrule
\end{tabular}%
}
\caption{Performance of RECOST based on different knowledge sources. RECOST (Wiki) and RECOST (Flan v2) refer to RECOST with Wikipedia and Flan v2 as the external knowledge source respectively.}
\label{tab:knowledge_source}
\end{table}

\paragraph{Different External Knowledge Types}
To find out how external knowledge affects the final performance within our RECOST, we further explore this by conducting two experiments: one for random truthful demonstrations and another for synthetic demonstrations. 

\begin{table}[h]
\centering
\resizebox{0.9\columnwidth}{!}{%
\begin{tabular}{@{}cccc@{}}
\toprule
\multirow{2}{*}{AlpacaEval} & \multicolumn{3}{c}{In-context Knowledge Type} \\ \cmidrule(l){2-4} 
                            & Synthetic       & Random       & Retrieved       \\ \midrule
helpful\_base               & 41.86           & 44.19        & 48.06        \\
koala                       & 31.41           & 37.18        & 41.03        \\
oasst                       & 38.30           & 40.96        & 41.49        \\
selfinstruct                & 28.29           & 19.52        & 28.57        \\
vicuna                      & 41.25           & 46.25        & 47.50        \\ \midrule
Overall                     & 34.76           & 34.70        & \textbf{39.19}        \\ \bottomrule
\end{tabular}%
}
\caption{Performance of RECOST with different in-context knowledge types on AlpacaEval benchmark.}
\label{tab:ick}
\end{table}

Specifically, synthetic in-context knowledge refers to retrieving similar demonstrations from the Alpaca dataset itself, while random in-context knowledge refers to utilizing random demonstrations as in-context knowledge. Table~\ref{tab:ick} shows the effects in dimensions of both truthfulness and similarity of in-context knowledge. The original RECOST outperforms the two mentioned in-context knowledge types on all subsets and the overall score of AlpacaEval benchmark.

Besides, we further explore the robustness of relative uncertainty as the number of retrieved demonstrations changes. In Section~\ref{sec:robustness}, we use the Jaccard similarity coefficient to compare the similarity of the top 10\% examples sorted by our proposed relative uncertainty with 1 to 5 demonstrations retrieved.



\subsubsection{Ablation on Diversity Sampling}
\paragraph{The effectiveness of diversity-consistent sampling.}To further evaluate the effectiveness of the diversity-consistent sampling module of RECOST, we compare RECOST under different settings. As presented in Table~\ref{tab:diversity_ablation}, with diversity-consistent sampling, RECOST achieves slight improvement on both benchmarks.
\begin{table}[h]
\resizebox{\columnwidth}{!}{%
\begin{tabular}{@{}c|c|cc@{}}
\toprule
Dataset                      & Setup                          & OpenLLM & AlpacaEval \\ \midrule
\multirow{2}{*}{Alpaca}
                             & w/o. diversity sampling & 56.05   & 37.80      \\
                             & w/. diversity sampling  & \textbf{56.07}   & \textbf{39.19}      \\ \midrule
\multirow{2}{*}{Alpaca-gpt4}
                             & w/o. diversity sampling & 59.38   & 62.80      \\
                             & w/. diversity sampling  & \textbf{59.60}   & \textbf{63.50}      \\ \bottomrule
\end{tabular}%
}
\caption{Ablation study on diversity sampling.}
\label{tab:diversity_ablation}
\end{table}

\paragraph{Weigh between uncertainty and diversity.}
Our experimental results also imply that the balance between uncertainty and diversity should not be neglected. Table~\ref{tab:diversity_uncertainty} gives results of RECOST under variant diversity. By appropriately adjusting tolerance, we can enhance performance. However, excessively high diversity can detrimentally impact efficiency. Here we simply use the mean cosine similarity as the diversity of the selected datasets, where lower mean cosine similarity leads to datasets with higher diversity.

\begin{table}[h]
\centering
\resizebox{\columnwidth}{!}{%
\begin{tabular}{@{}c|ccccc@{}}
\toprule
Tolerance   & -      & 468    & 104    & 52     & 26     \\ \midrule
Mean CosSim & 0.6852 & 0.6840 & 0.6815 & 0.6805 & 0.6794 \\ \midrule
AlpacaEval  & 37.80  & 39.19  & 38.14  & 38.57  & 36.34  \\ \bottomrule
\end{tabular}%
}
\caption{Performance of RECOST with increasing diversity. Tolerance `-' refers to RECOST without diversity sampling.}
\label{tab:diversity_uncertainty}
\end{table}

\subsection{Analysis}
\subsubsection{Effectiveness of RECOST}
\begin{figure}[h]
\centering
\includegraphics[width=\columnwidth]{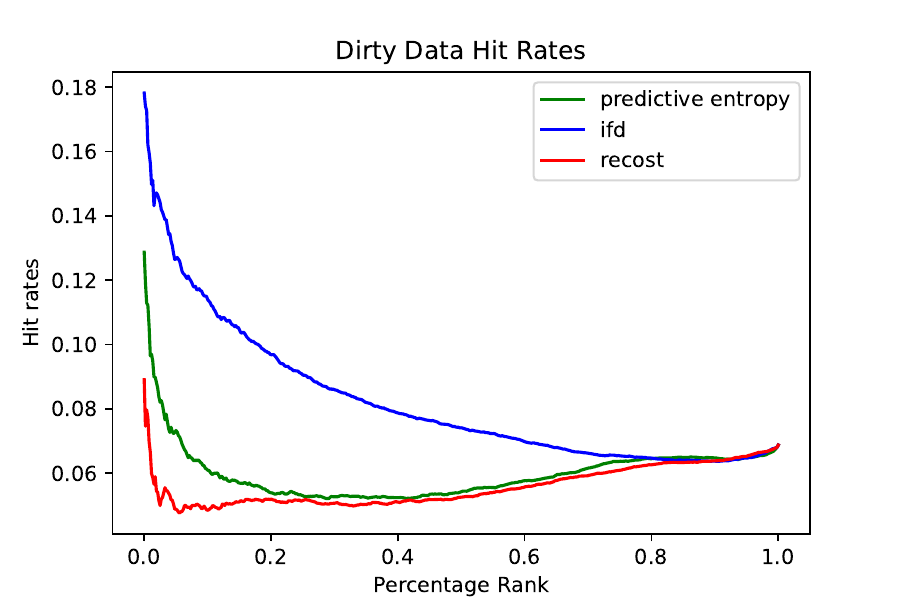}
\caption{Comparison of dirty data hit rates under different metrics on the Alpaca dataset calculated by LLaMA-2-7b. The horizontal axis represents the percentage ranking, while the vertical axis denotes the proportion of corrupted data within the data preceding that percentage threshold. Given the number of dirty data in the top $i$ data points as $d_i$, the hit rate at $i$ is calculated by $d_i/i$. The dirty data is collected by comparing Alpaca with Alpaca-cleaned.}
\label{fig:compare_hit}
\end{figure}

Moreover, we make a comparison of dirty data hit rates on RECOST, predictive entropy, and IFD. Similar to Figure~\ref{fig:pe_dirty_hit_rates}, we calculate the hit rates of three metrics. As presented in Figure~\ref{fig:compare_hit}, our proposed RECOST outperforms both metrics remarkably on top 20\% samples. Besides, we further compare the performance of the three metrics above in Section~\ref{sec:effcts}.

\subsubsection{The Effects of Mixed Weight}
To find the optimal mixed weight $w$, we conduct extensive experiments with different weights on both datasets. We set the mixed weight $w$ to 0, 0.25, 0.5, 0.75, and 1. In particular, $w=1$ refers to vanilla predictive entropy while $w=0$ refers to our proposed relative predictive entropy.

\begin{figure}[h]
\centering
\includegraphics[width=0.9\columnwidth]{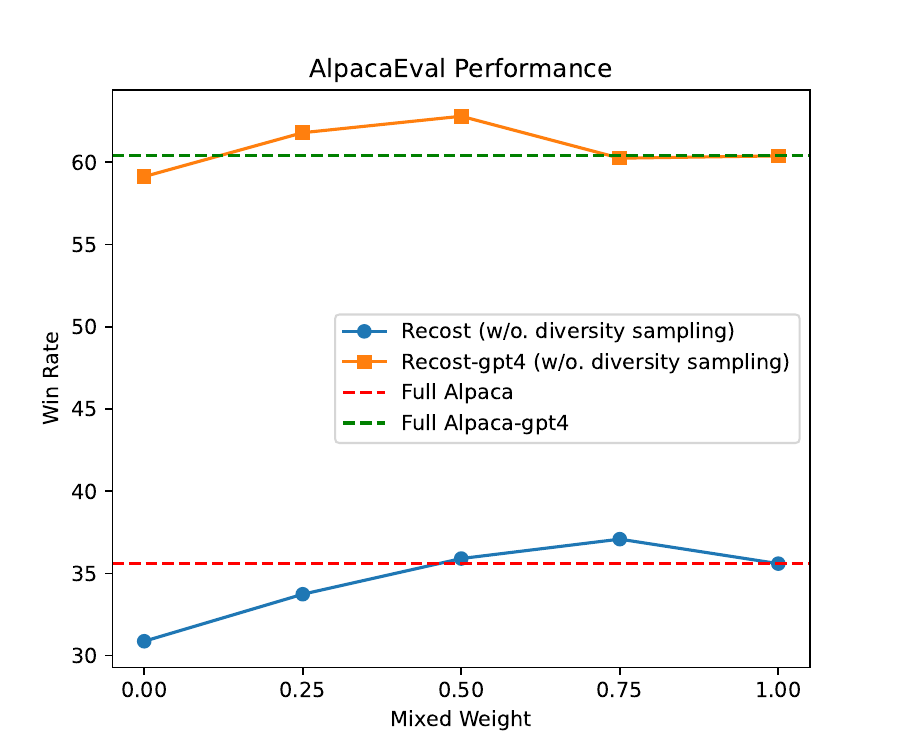}
\caption{Performance of relative predictive entropy as the change of mixed weight $w$.}
\label{fig:mixed_weight}
\end{figure}

As presented in Figure~\ref{fig:mixed_weight}, the performance of the models on AlpacaEval demonstrates an initial increase followed by a decline, peaking at values of $w=0.5$ and $w=0.75$, respectively. Specifically, due to the overall inferior quality of the Alpaca dataset in comparison to Alpaca-gpt4, coupled with its shorter average response length, a greater emphasis on weighting is inclined towards selecting longer samples, which in turn yields superior outcomes.

\section{Conclusion}
In this paper, we demonstrate \textbf{RECOST}, an effective method to select high-quality instruction data from synthetic instruction datasets with the help of external knowledge. Our method outperforms the fully-trained models with only 1\% training data, which surpasses the previous methods under the same settings. Overall, our approach underscores the significance and efficacy of integrating our proposed relative uncertainty into data-efficient instruction tuning for synthetic datasets, providing a viable avenue for this field.

\section{Limitations}
The primary limitation of our work lies in the necessity of incorporating additional external knowledge. However, thanks to the development of traditional NLP tasks during the pre-LLM era and the current advancements in retrieval-based ICL, we can easily obtain a vast amount of authentic and reliable external knowledge to meet our requirements. Overall, our research empirically validates the feasibility of integrating exogenous knowledge in the data filtering process based on synthetic data. Although this introduces a minor overhead in data preprocessing, it significantly outperforms previous methods within an acceptable cost margin, offering new perspectives in the realm of data efficiency.

\bibliography{anthology}

\begin{thebibliography}{32}
\expandafter\ifx\csname natexlab\endcsname\relax\def\natexlab#1{#1}\fi

\bibitem[{Brown et~al.(2020)Brown, Mann, Ryder, Subbiah, Kaplan, Dhariwal, Neelakantan, Shyam, Sastry, Askell et~al.}]{brown2020language}
Tom Brown, Benjamin Mann, Nick Ryder, Melanie Subbiah, Jared~D Kaplan, Prafulla Dhariwal, Arvind Neelakantan, Pranav Shyam, Girish Sastry, Amanda Askell, et~al. 2020.
\newblock Language models are few-shot learners.
\newblock \emph{Advances in neural information processing systems}, 33:1877--1901.

\bibitem[{Cao et~al.(2023)Cao, Kang, Wang, and Sun}]{cao2023instruction}
Yihan Cao, Yanbin Kang, Chi Wang, and Lichao Sun. 2023.
\newblock \href {http://arxiv.org/abs/2307.06290} {Instruction mining: When data mining meets large language model finetuning}.

\bibitem[{Chen et~al.(2023{\natexlab{a}})Chen, Zhang, Zhang, Yang, Hu, Ma, Yanggong, and Zhao}]{chen2023maybe}
Hao Chen, Yiming Zhang, Qi~Zhang, Hantao Yang, Xiaomeng Hu, Xuetao Ma, Yifan Yanggong, and Junbo Zhao. 2023{\natexlab{a}}.
\newblock \href {http://arxiv.org/abs/2305.09246} {Maybe only 0.5\% data is needed: A preliminary exploration of low training data instruction tuning}.

\bibitem[{Chen et~al.(2023{\natexlab{b}})Chen, Li, Yan, Wang, Gunaratna, Yadav, Tang, Srinivasan, Zhou, Huang, and Jin}]{chen2023alpagasus}
Lichang Chen, Shiyang Li, Jun Yan, Hai Wang, Kalpa Gunaratna, Vikas Yadav, Zheng Tang, Vijay Srinivasan, Tianyi Zhou, Heng Huang, and Hongxia Jin. 2023{\natexlab{b}}.
\newblock \href {http://arxiv.org/abs/2307.08701} {Alpagasus: Training a better alpaca with fewer data}.

\bibitem[{Chung et~al.(2022)Chung, Hou, Longpre, Zoph, Tay, Fedus, Li, Wang, Dehghani, Brahma et~al.}]{chung2022scaling}
Hyung~Won Chung, Le~Hou, Shayne Longpre, Barret Zoph, Yi~Tay, William Fedus, Yunxuan Li, Xuezhi Wang, Mostafa Dehghani, Siddhartha Brahma, et~al. 2022.
\newblock Scaling instruction-finetuned language models.
\newblock \emph{arXiv preprint arXiv:2210.11416}.

\bibitem[{Duan et~al.(2023)Duan, Cheng, Wang, Zavalny, Wang, Xu, Kailkhura, and Xu}]{duan2023shifting}
Jinhao Duan, Hao Cheng, Shiqi Wang, Alex Zavalny, Chenan Wang, Renjing Xu, Bhavya Kailkhura, and Kaidi Xu. 2023.
\newblock \href {http://arxiv.org/abs/2307.01379} {Shifting attention to relevance: Towards the uncertainty estimation of large language models}.

\bibitem[{Gao et~al.(2023)Gao, Tow, Abbasi, Biderman, Black, DiPofi, Foster, Golding, Hsu, Le~Noac'h, Li, McDonell, Muennighoff, Ociepa, Phang, Reynolds, Schoelkopf, Skowron, Sutawika, Tang, Thite, Wang, Wang, and Zou}]{eval-harness}
Leo Gao, Jonathan Tow, Baber Abbasi, Stella Biderman, Sid Black, Anthony DiPofi, Charles Foster, Laurence Golding, Jeffrey Hsu, Alain Le~Noac'h, Haonan Li, Kyle McDonell, Niklas Muennighoff, Chris Ociepa, Jason Phang, Laria Reynolds, Hailey Schoelkopf, Aviya Skowron, Lintang Sutawika, Eric Tang, Anish Thite, Ben Wang, Kevin Wang, and Andy Zou. 2023.
\newblock \href {https://doi.org/10.5281/zenodo.10256836} {A framework for few-shot language model evaluation}.

\bibitem[{Hendrycks et~al.(2021)Hendrycks, Burns, Basart, Zou, Mazeika, Song, and Steinhardt}]{hendrycks2021measuring}
Dan Hendrycks, Collin Burns, Steven Basart, Andy Zou, Mantas Mazeika, Dawn Song, and Jacob Steinhardt. 2021.
\newblock \href {http://arxiv.org/abs/2009.03300} {Measuring massive multitask language understanding}.

\bibitem[{Kadavath et~al.(2022)Kadavath, Conerly, Askell, Henighan, Drain, Perez, Schiefer, Hatfield-Dodds, DasSarma, Tran-Johnson, Johnston, El-Showk, Jones, Elhage, Hume, Chen, Bai, Bowman, Fort, Ganguli, Hernandez, Jacobson, Kernion, Kravec, Lovitt, Ndousse, Olsson, Ringer, Amodei, Brown, Clark, Joseph, Mann, McCandlish, Olah, and Kaplan}]{kadavath2022language}
Saurav Kadavath, Tom Conerly, Amanda Askell, Tom Henighan, Dawn Drain, Ethan Perez, Nicholas Schiefer, Zac Hatfield-Dodds, Nova DasSarma, Eli Tran-Johnson, Scott Johnston, Sheer El-Showk, Andy Jones, Nelson Elhage, Tristan Hume, Anna Chen, Yuntao Bai, Sam Bowman, Stanislav Fort, Deep Ganguli, Danny Hernandez, Josh Jacobson, Jackson Kernion, Shauna Kravec, Liane Lovitt, Kamal Ndousse, Catherine Olsson, Sam Ringer, Dario Amodei, Tom Brown, Jack Clark, Nicholas Joseph, Ben Mann, Sam McCandlish, Chris Olah, and Jared Kaplan. 2022.
\newblock \href {http://arxiv.org/abs/2207.05221} {Language models (mostly) know what they know}.

\bibitem[{Li et~al.(2023{\natexlab{a}})Li, Zhang, Li, Chen, Chen, Cheng, Wang, Zhou, and Xiao}]{Li2023FromQT}
Ming Li, Yong Zhang, Zhitao Li, Jiuhai Chen, Lichang Chen, Ning Cheng, Jianzong Wang, Tianyi Zhou, and Jing Xiao. 2023{\natexlab{a}}.
\newblock \href {https://api.semanticscholar.org/CorpusID:261076515} {From quantity to quality: Boosting llm performance with self-guided data selection for instruction tuning}.
\newblock \emph{ArXiv}, abs/2308.12032.

\bibitem[{Li et~al.(2023{\natexlab{b}})Li, Zhang, Dubois, Taori, Gulrajani, Guestrin, Liang, and Hashimoto}]{alpaca_eval}
Xuechen Li, Tianyi Zhang, Yann Dubois, Rohan Taori, Ishaan Gulrajani, Carlos Guestrin, Percy Liang, and Tatsunori~B. Hashimoto. 2023{\natexlab{b}}.
\newblock Alpacaeval: An automatic evaluator of instruction-following models.
\newblock \url{https://github.com/tatsu-lab/alpaca_eval}.

\bibitem[{Li et~al.(2023{\natexlab{c}})Li, Hui, Xia, Yang, Yang, Zhang, Si, Liu, Liu, Huang et~al.}]{li2023one}
Yunshui Li, Binyuan Hui, Xiaobo Xia, Jiaxi Yang, Min Yang, Lei Zhang, Shuzheng Si, Junhao Liu, Tongliang Liu, Fei Huang, et~al. 2023{\natexlab{c}}.
\newblock One shot learning as instruction data prospector for large language models.
\newblock \emph{arXiv preprint arXiv:2312.10302}.

\bibitem[{Lin et~al.(2022)Lin, Hilton, and Evans}]{lin2022truthfulqa}
Stephanie Lin, Jacob Hilton, and Owain Evans. 2022.
\newblock \href {http://arxiv.org/abs/2109.07958} {Truthfulqa: Measuring how models mimic human falsehoods}.

\bibitem[{Liu et~al.(2024)Liu, Zeng, He, Jiang, and He}]{liu2024what}
Wei Liu, Weihao Zeng, Keqing He, Yong Jiang, and Junxian He. 2024.
\newblock \href {https://openreview.net/forum?id=BTKAeLqLMw} {What makes good data for alignment? a comprehensive study of automatic data selection in instruction tuning}.
\newblock In \emph{The Twelfth International Conference on Learning Representations}.

\bibitem[{Longpre et~al.(2023)Longpre, Hou, Vu, Webson, Chung, Tay, Zhou, Le, Zoph, Wei et~al.}]{longpre2023flan}
Shayne Longpre, Le~Hou, Tu~Vu, Albert Webson, Hyung~Won Chung, Yi~Tay, Denny Zhou, Quoc~V Le, Barret Zoph, Jason Wei, et~al. 2023.
\newblock The flan collection: Designing data and methods for effective instruction tuning.
\newblock \emph{arXiv preprint arXiv:2301.13688}.

\bibitem[{Luo et~al.(2023{\natexlab{a}})Luo, Sun, Xu, Zhao, Lou, Tao, Geng, Lin, Chen, and Zhang}]{luo2023wizardmath}
Haipeng Luo, Qingfeng Sun, Can Xu, Pu~Zhao, Jianguang Lou, Chongyang Tao, Xiubo Geng, Qingwei Lin, Shifeng Chen, and Dongmei Zhang. 2023{\natexlab{a}}.
\newblock Wizardmath: Empowering mathematical reasoning for large language models via reinforced evol-instruct.
\newblock \emph{arXiv preprint arXiv:2308.09583}.

\bibitem[{Luo et~al.(2024)Luo, Xu, Liu, Pasupat, and Kazemi}]{luo2024incontext}
Man Luo, Xin Xu, Yue Liu, Panupong Pasupat, and Mehran Kazemi. 2024.
\newblock \href {http://arxiv.org/abs/2401.11624} {In-context learning with retrieved demonstrations for language models: A survey}.

\bibitem[{Luo et~al.(2023{\natexlab{b}})Luo, Xu, Zhao, Sun, Geng, Hu, Tao, Ma, Lin, and Jiang}]{luo2023wizardcoder}
Ziyang Luo, Can Xu, Pu~Zhao, Qingfeng Sun, Xiubo Geng, Wenxiang Hu, Chongyang Tao, Jing Ma, Qingwei Lin, and Daxin Jiang. 2023{\natexlab{b}}.
\newblock Wizardcoder: Empowering code large language models with evol-instruct.
\newblock \emph{arXiv preprint arXiv:2306.08568}.

\bibitem[{Ouyang et~al.(2022)Ouyang, Wu, Jiang, Almeida, Wainwright, Mishkin, Zhang, Agarwal, Slama, Ray et~al.}]{ouyang2022training}
Long Ouyang, Jeffrey Wu, Xu~Jiang, Diogo Almeida, Carroll Wainwright, Pamela Mishkin, Chong Zhang, Sandhini Agarwal, Katarina Slama, Alex Ray, et~al. 2022.
\newblock Training language models to follow instructions with human feedback.
\newblock \emph{Advances in Neural Information Processing Systems}, 35:27730--27744.

\bibitem[{Peng et~al.(2023)Peng, Li, He, Galley, and Gao}]{peng2023instruction}
Baolin Peng, Chunyuan Li, Pengcheng He, Michel Galley, and Jianfeng Gao. 2023.
\newblock Instruction tuning with gpt-4.
\newblock \emph{arXiv preprint arXiv:2304.03277}.

\bibitem[{Rubin et~al.(2021)Rubin, Herzig, and Berant}]{rubin2021learning}
Ohad Rubin, Jonathan Herzig, and Jonathan Berant. 2021.
\newblock Learning to retrieve prompts for in-context learning.
\newblock \emph{arXiv preprint arXiv:2112.08633}.

\bibitem[{Sener and Savarese(2018)}]{sener2018active}
Ozan Sener and Silvio Savarese. 2018.
\newblock \href {https://openreview.net/forum?id=H1aIuk-RW} {Active learning for convolutional neural networks: A core-set approach}.
\newblock In \emph{International Conference on Learning Representations}.

\bibitem[{Taori et~al.(2023)Taori, Gulrajani, Zhang, Dubois, Li, Guestrin, Liang, and Hashimoto}]{alpaca}
Rohan Taori, Ishaan Gulrajani, Tianyi Zhang, Yann Dubois, Xuechen Li, Carlos Guestrin, Percy Liang, and Tatsunori~B. Hashimoto. 2023.
\newblock Stanford alpaca: An instruction-following llama model.
\newblock \url{https://github.com/tatsu-lab/stanford_alpaca}.

\bibitem[{Wang et~al.(2024)Wang, Yang, and Wei}]{wang2024learning}
Liang Wang, Nan Yang, and Furu Wei. 2024.
\newblock \href {http://arxiv.org/abs/2307.07164} {Learning to retrieve in-context examples for large language models}.

\bibitem[{Wang et~al.(2023)Wang, Ivison, Dasigi, Hessel, Khot, Chandu, Wadden, MacMillan, Smith, Beltagy, and Hajishirzi}]{wang2023far}
Yizhong Wang, Hamish Ivison, Pradeep Dasigi, Jack Hessel, Tushar Khot, Khyathi~Raghavi Chandu, David Wadden, Kelsey MacMillan, Noah~A. Smith, Iz~Beltagy, and Hannaneh Hajishirzi. 2023.
\newblock \href {http://arxiv.org/abs/2306.04751} {How far can camels go? exploring the state of instruction tuning on open resources}.

\bibitem[{Wang et~al.(2022)Wang, Kordi, Mishra, Liu, Smith, Khashabi, and Hajishirzi}]{selfinstruct}
Yizhong Wang, Yeganeh Kordi, Swaroop Mishra, Alisa Liu, Noah~A. Smith, Daniel Khashabi, and Hannaneh Hajishirzi. 2022.
\newblock Self-instruct: Aligning language model with self generated instructions.

\bibitem[{Wei et~al.()Wei, Bosma, Zhao, Guu, Yu, Lester, Du, Dai, and Le}]{weifinetuned}
Jason Wei, Maarten Bosma, Vincent Zhao, Kelvin Guu, Adams~Wei Yu, Brian Lester, Nan Du, Andrew~M Dai, and Quoc~V Le.
\newblock Finetuned language models are zero-shot learners.
\newblock In \emph{International Conference on Learning Representations}.

\bibitem[{Xu et~al.(2023)Xu, Sun, Zheng, Geng, Zhao, Feng, Tao, and Jiang}]{xu2023wizardlm}
Can Xu, Qingfeng Sun, Kai Zheng, Xiubo Geng, Pu~Zhao, Jiazhan Feng, Chongyang Tao, and Daxin Jiang. 2023.
\newblock Wizardlm: Empowering large language models to follow complex instructions.
\newblock \emph{arXiv preprint arXiv:2304.12244}.

\bibitem[{Yadav et~al.(2019)Yadav, Bethard, and Surdeanu}]{Yadav_2019}
Vikas Yadav, Steven Bethard, and Mihai Surdeanu. 2019.
\newblock \href {https://doi.org/10.18653/v1/d19-1260} {Quick and (not so) dirty: Unsupervised selection of justification sentences for multi-hop question answering}.
\newblock In \emph{Proceedings of the 2019 Conference on Empirical Methods in Natural Language Processing and the 9th International Joint Conference on Natural Language Processing (EMNLP-IJCNLP)}. Association for Computational Linguistics.

\bibitem[{Zellers et~al.(2019)Zellers, Holtzman, Bisk, Farhadi, and Choi}]{zellers2019hellaswag}
Rowan Zellers, Ari Holtzman, Yonatan Bisk, Ali Farhadi, and Yejin Choi. 2019.
\newblock \href {http://arxiv.org/abs/1905.07830} {Hellaswag: Can a machine really finish your sentence?}

\bibitem[{Zhang et~al.(2023)Zhang, Xiao, Liu, Dou, and Nie}]{llm_embedder}
Peitian Zhang, Shitao Xiao, Zheng Liu, Zhicheng Dou, and Jian-Yun Nie. 2023.
\newblock \href {http://arxiv.org/abs/2310.07554} {Retrieve anything to augment large language models}.

\bibitem[{Zhou et~al.(2023)Zhou, Liu, Xu, Iyer, Sun, Mao, Ma, Efrat, Yu, Yu, Zhang, Ghosh, Lewis, Zettlemoyer, and Levy}]{zhou2023lima}
Chunting Zhou, Pengfei Liu, Puxin Xu, Srini Iyer, Jiao Sun, Yuning Mao, Xuezhe Ma, Avia Efrat, Ping Yu, Lili Yu, Susan Zhang, Gargi Ghosh, Mike Lewis, Luke Zettlemoyer, and Omer Levy. 2023.
\newblock \href {http://arxiv.org/abs/2305.11206} {Lima: Less is more for alignment}.

\end{thebibliography}

\newpage
\appendix
\section{Appendix}
\label{sec:appendix}


\subsection{Robustness Analysis of Relative Uncertainty}
\label{sec:robustness}
\begin{figure}[h]
\centering
\begin{subfigure}[b]{\columnwidth}
\centering
\includegraphics[width=\columnwidth]{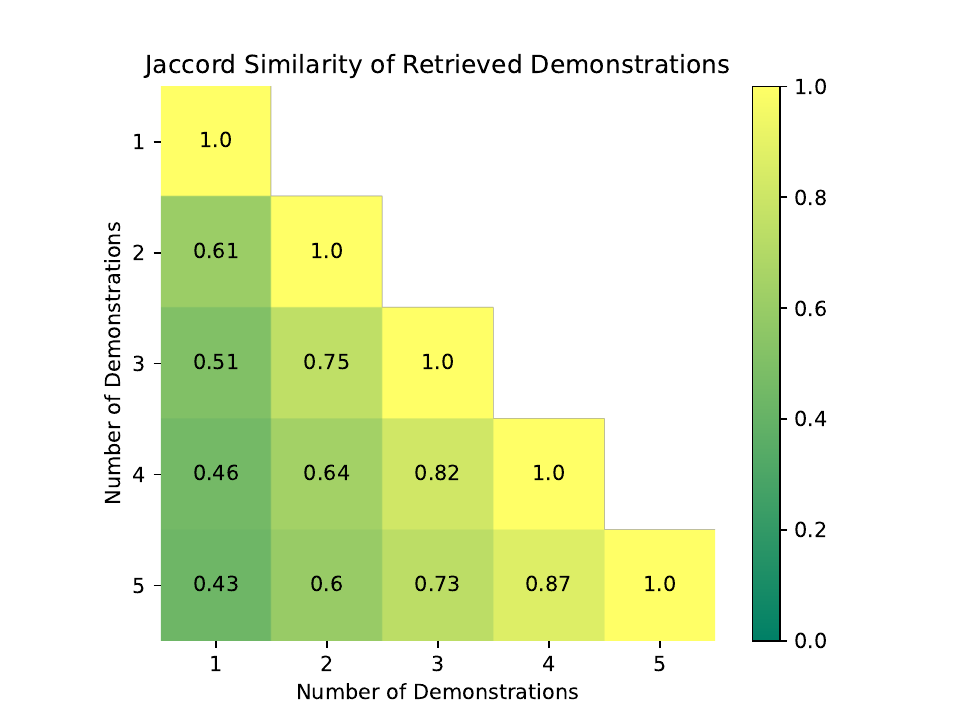}
\caption{}
\label{fig:jsc_retrieved}
\end{subfigure}
\begin{subfigure}[b]{\columnwidth}
\centering
\includegraphics[width=\columnwidth]{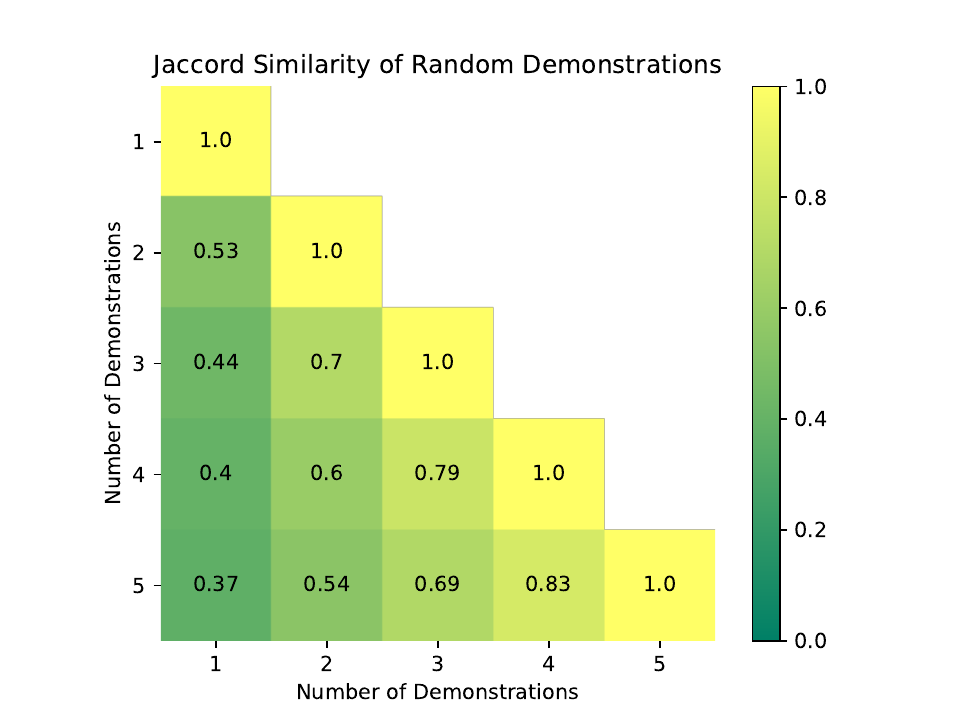}
\caption{}
\label{fig:jsc_random}
\end{subfigure}
\caption{The robustness of relative uncertainty as the number of demonstrations changed when we used random samples as demonstrations. In Figure~\ref{fig:jsc_retrieved}, we use retrieved demonstrations, while we use random demonstrations in Figure~\ref{fig:jsc_random}}.
\label{fig:retrieval-vs-random}
\end{figure}

As presented in \ref{fig:retrieval-vs-random}, with the increase in the number of demonstrations, RECOST with retrieved demonstrations tends to be more stable than that with random demonstrations.

\newpage
\subsection{Effectiveness of RECOST}
\label{sec:effcts}
We present the comparison between the aforementioned metrics in Table~\ref{tab:comp_hits}. RECOST outperforms all baselines in both benchmarks.

\begin{table}[h]
\centering
\resizebox{0.9\columnwidth}{!}{%
\begin{tabular}{@{}c|c|cc@{}}
\toprule
Method               & Data size & OpenLLM & AlpacaEval \\ \midrule
Full Alpaca          & 52002     & 55.06   & 27.75      \\
IFD                  & 3111      & 55.78   & 36.78      \\
PE & 520       & 55.31   & 35.59      \\
RECOST               & 520       & \textbf{56.07}   & \textbf{39.19}     \\ \bottomrule
\end{tabular}%
}
\caption{Comparison of benchmark scores under different metrics on the Alpaca dataset.}
\label{tab:comp_hits}
\end{table}



\end{document}